\documentclass[a4paper,fleqn]{cas-sc}
\usepackage{amsmath}
\usepackage{dsfont}
\usepackage[authoryear,longnamesfirst]{natbib}
\usepackage{threeparttable}
\usepackage{algorithmic}  
\usepackage{algorithm} 
\usepackage[algo2e]{algorithm2e}
\usepackage{graphicx}

\usepackage{booktabs}
\usepackage{makecell}
\def\tsc#1{\csdef{#1}{\textsc{\lowercase{#1}}\xspace}}
\tsc{WGM}
\tsc{QE}
\tsc{EP}
\tsc{PMS}
\tsc{BEC}
\tsc{DE}
\begin{document}
\let\WriteBookmarks\relax
\def\floatpagepagefraction{1}
\def\textpagefraction{.001}

\shorttitle{Preprint}

\shortauthors{Yao et~al.}

\title [mode = title]{DiffRaman: A Conditional Latent Denoising Diffusion Probabilistic Model for Bacterial Raman Spectroscopy Identification Under Limited Data Conditions}                      


\affiliation[1]{organization={
State Key Laboratory of Precision Measurement Technology and Instruments, Tsinghua University}, 
    city={Beijing},
    postcode={100084}, 
    country={China}}

\author[1]{Haiming Yao}[style=chinese, orcid=0000-0003-1419-5489
]
\ead{yhm22@mails.tsinghua.edu.cn}

\author[1]{Wei Luo}[style=chinese]
\ead{luow23@mails.tsinghua.edu.cn}

\author[1]{Ang Gao}[style=chinese]
\ead{ga24@mails.tsinghua.edu.cn}

\author[1]{Tao Zhou}[style=chinese]
\ead{zhoutao24@mails.tsinghua.edu.cn}

\author[1]{Xue Wang}[style=chinese, orcid=0000-0003-4842-3160]
\cormark[1]
\ead{wangxue@mail.tsinghua.edu.cn}

\cortext[cor1]{Corresponding author}

\begin{abstract}
Raman spectroscopy has attracted significant attention in various biochemical detection fields, especially in the rapid identification of pathogenic bacteria. The integration of this technology with deep learning to facilitate automated bacterial Raman spectroscopy diagnosis has emerged as a key focus in recent research. However, the diagnostic performance of existing deep learning methods largely depends on a sufficient dataset, and in scenarios where there is a limited availability of Raman spectroscopy data, it is inadequate to fully optimize the numerous parameters of deep neural networks. To address these challenges, this paper proposes a data generation method utilizing deep generative models to expand the data volume and enhance the recognition accuracy of bacterial Raman spectra. Specifically, we introduce DiffRaman, a conditional latent denoising diffusion probability model for Raman spectra generation. Our approach begins with applying a two-dimensional figure transformation to the Raman spectral data. Following this, we utilize the encoder of a Vector Quantized Variational Autoencoder (VQ-VAE) to compress the Raman figure into a lower-dimensional latent space. We then construct a Conditional Denoising Diffusion Probabilistic Model (DDPM) for representation learning and data augmentation. Ultimately, the decoder of the VQ-VAE is employed to reconstruct the spectrum from its low-dimensional latent representation. Experimental results demonstrate that synthetic bacterial Raman spectra generated by DiffRaman can effectively emulate real experimental spectra, thereby enhancing the performance of diagnostic models, especially under conditions of limited data. Furthermore, compared to existing generative models, the proposed DiffRaman offers improvements in both generation quality and computational efficiency. Our DiffRaman approach offers a well-suited solution for automated bacteria Raman spectroscopy diagnosis in data-scarce scenarios, offering new insights into alleviating the labor of spectroscopic measurements and enhancing rare bacteria identification.

\end{abstract}

\begin{highlights}
\item This paper proposed to employ data generation to enhance Raman spectra identification under limited data.
\item A new generative model-based Raman spectra generation method is proposed.
\item A conditional latent denoising diffusion probability model is developed to generate synthetic Raman spectra.
\end{highlights}

% Keywords

\maketitle

\section{Introduction}
\label{sec:introduction}

Bacterial infections continue to be a major cause of global death, making the identification of pathogenic bacteria and their susceptibility to antibiotics vital in clinical applications. This is essential for accurately diagnosing diseases caused by bacterial infections and determining treatment strategies. Raman spectroscopy, leveraging molecular vibrations, serves as a potent analytical tool capable of providing intricate molecular fingerprint information, thereby facilitating the analysis of diverse biochemical characteristics. In recent years, its non-destructive, label-free, and rapid detection characteristics have garnered widespread favor, leading to extensive application in the field of bacterial identification \cite{r1,r67}. The biochemical phenotypes of various subjects yield distinct spectral fingerprint in their respective Raman spectral data. However, in practical applications, the intricate composition of spectral data poses a significant challenge for direct and accurate identification by the human eye. Furthermore, interpreting the biochemical significance associated with different wavelengths necessitates extensive expert knowledge\cite{r1}. Consequently, the automated analysis of spectral data has emerged as a critical concern\cite{r1}\cite{r40}\cite{r41}.

Various methods have been proposed to effectively analyze Raman spectroscopy signals and achieve their identification. The Ramanome method proposed in \cite{r42} utilizes specific wavelengths to conduct manual spectral analysis. However, the applicability of this expert knowledge-based approach is confined to particular tasks and lacks generalizability to other contexts. Another potential route involves employing data-driven machine learning methods. Conventional machine learning methods primarily involve extracting signal features pertinent to domain knowledge, followed by the classification of these extracted features using appropriate classifiers. For instance, in \cite{r43}, a partial least squares discriminant analysis method was proposed for the classification of COVID-19 utilizing Raman spectroscopy. In \cite{r44}\cite{r45}, principal component analysis (PCA) is employed for feature extraction, while support vector machines (SVM) are utilized as classifiers to categorize diseases based on Raman spectra. In existing research, other frequently employed methods also include k-nearest neighbors (KNN)\cite{r46}, discriminant function analysis (DFA)\cite{r47}, and random forests\cite{r48}. Although numerous methods and their variants have been developed, this technical approach still has certain limitations: manually extracted features may loss implicit spectral characteristics, and the constructed classifiers are susceptible to overfitting, noise interference, and other issues.

Owing to automated representation extraction, deep learning has been extensively applied across various fields. In Raman spectroscopy analysis, it has already demonstrated the  capability to surpass the traditional machine learning methods in terms of accuracy, robustness, and generalizability. Various deep neural networks, including convolutional neural networks (CNNs)\cite{r1}, and recurrent neural networks (RNNs)\cite{r49}, have been utilized in Raman spectroscopy analysis. In \cite{r1}, transfer learning-enabled CNN was employed for large-scale bacterial identification. This research has been developed upon in subsequent studies, such as SANet\cite{r16} and U-Net\cite{r50}. In \cite{r51}, Raman spectroscopy and deep neural networks were employed for the in vitro and intraoperative pathological diagnosis of liver cancer. In \cite{r52}, the authors introduced a two-dimensional (2D) Raman figure technique combined with CNN for tumor diagnosis.

% However, due to the extensive number of parameters in deep neural networks, the optimization process to achieve an optimal solution often necessitates a substantial number of diverse training samples. Generally, acquiring this data requires significant human labor, as the collection process must be conducted in a strictly controlled experimental environment. Additionally, accurate labeling demands further labor and specialized expertise. Consequently, the limited availability of training spectra data is a significant impediment to the application of diagnostic systems based on deep neural networks in clinical settings. 

However, due to the extensive number of learnable parameters in deep neural networks, the optimization process for diagnostic models used in bacterial Raman spectroscopy often requires a large and diverse set of bacterial spectral samples to achieve robust performance. Nevertheless, obtaining a substantial amount of bacterial Raman spectroscopy data is not always feasible, such as in the single-cell measurement approach \cite{r15}, where collecting a large number of measurements would entail significant labor. Furthermore, as pointed out in \cite{r1}, in clinical application, each patient may only provide a minimal number of bacterial spectra for diagnosis. Additionally, accurately labeling and organizing a large amount of spectral data also requires more manpower and expertise. Therefore, developing a deep neural network-based diagnostic system with limited available bacterial Raman spectroscopy data is an important issue.

To address the issue of data insufficiency, researchers typically employ data augmentation techniques to generate additional data for expansion. For instance, Gaussian noise and various perturbations \cite{r16} are applied to the original Raman spectroscopy data. However, the benefits of such generated data are limited, as this approach primarily serves as a regularization method to prevent model overfitting. A more advanced approach involves employing generative models to produce synthetic data that is similar to the real data based on its distribution. Currently, widely used generative models include variational autoencoders (VAEs)\cite{r53} and generative adversarial networks (GANs)\cite{r54}. Among these models, the quality of samples generated by VAEs is limited by Gaussian prior assumptions, whereas the superior data generation capability of GANs has made them a popular solution for addressing data scarcity. For instance, in the field of industrial failure diagnosis, \cite{r55} discussed the issue of data scarcity by incorporating GANs. Despite the significant success of GANs, these models are generally regarded as challenging to train and are susceptible to mode collapse due to their adversarial training mechanism\cite{r56}.

Given the limitations of the aforementioned generative models, we introduce the physically inspired denoising diffusion probabilistic model (DDPM)\cite{r57} as an innovative approach for the generation of Raman spectra. This model demonstrates exceptional performance and training stability across various computer vision tasks, including image generation\cite{r59} and image editing\cite{r58}. Specifically, DDPM generates new samples that follow the distribution of the original data through a parameterized Markov process using variational inference. This process employs noise as a medium, treating the chain of noise as a learnable variable. The variable is iteratively degraded until the noise is completely removed, thus restoring the original input or producing synthetic data with accurate statistical characteristics and patterns\cite{r57}.

In this paper, we propose the DiffRaman, a conditional latent denoising diffusion probabilistic model to synthesize additional Raman spectroscopy data, thereby enhancing the spectral diagnostic accuracy under limited data situations. The original DDPM operates in the data space, requiring substantial memory resources for optimization and inference \cite{r60}. To reduce computational load while maintaining generation performance, we introduced an autoencoder to compress the original spectral data into a low-dimensional latent space. We further employed a vector quantization variational autoencoder (VQ-VAE)\cite{r61} to enhance the autoencoder's performance. Following this, a diffusion process is executed on the low-dimensional data within the latent discrete space of the VQ-VAE, incorporating a conditional mechanism to regulate the diffusion process and retain the category semantics of the generated data. Finally, the generated low-dimensional data is decoded to produce the new spectrum. To the best of our knowledge, our work represents the initial endeavor to introduce the concept of diffusion generation within the Raman spectroscopy diagnostics community. The primary contributions of this work can be summarized as follows.

\begin{enumerate}
\item We introduced the concept of employing data generation methods in the realm of Raman spectroscopy diagnostics to expand data capacity and enhance the accuracy of diagnostic models with limited data.

\item We proposed a novel DiffRaman model and introduced DDPM for the first time in the field of Raman spectroscopy for spectral generation.

\item Extensive experimental analyses conducted on large-scale bacterial Raman spectroscopy datasets indicate that the proposed DiffRaman method enhances diagnostic accuracy with limited data, exhibiting superior performance compared to existing approaches and underscoring its potential for clinical application.

\end{enumerate}

\begin{figure}[!t]
\centerline{\includegraphics[width=15.0cm]{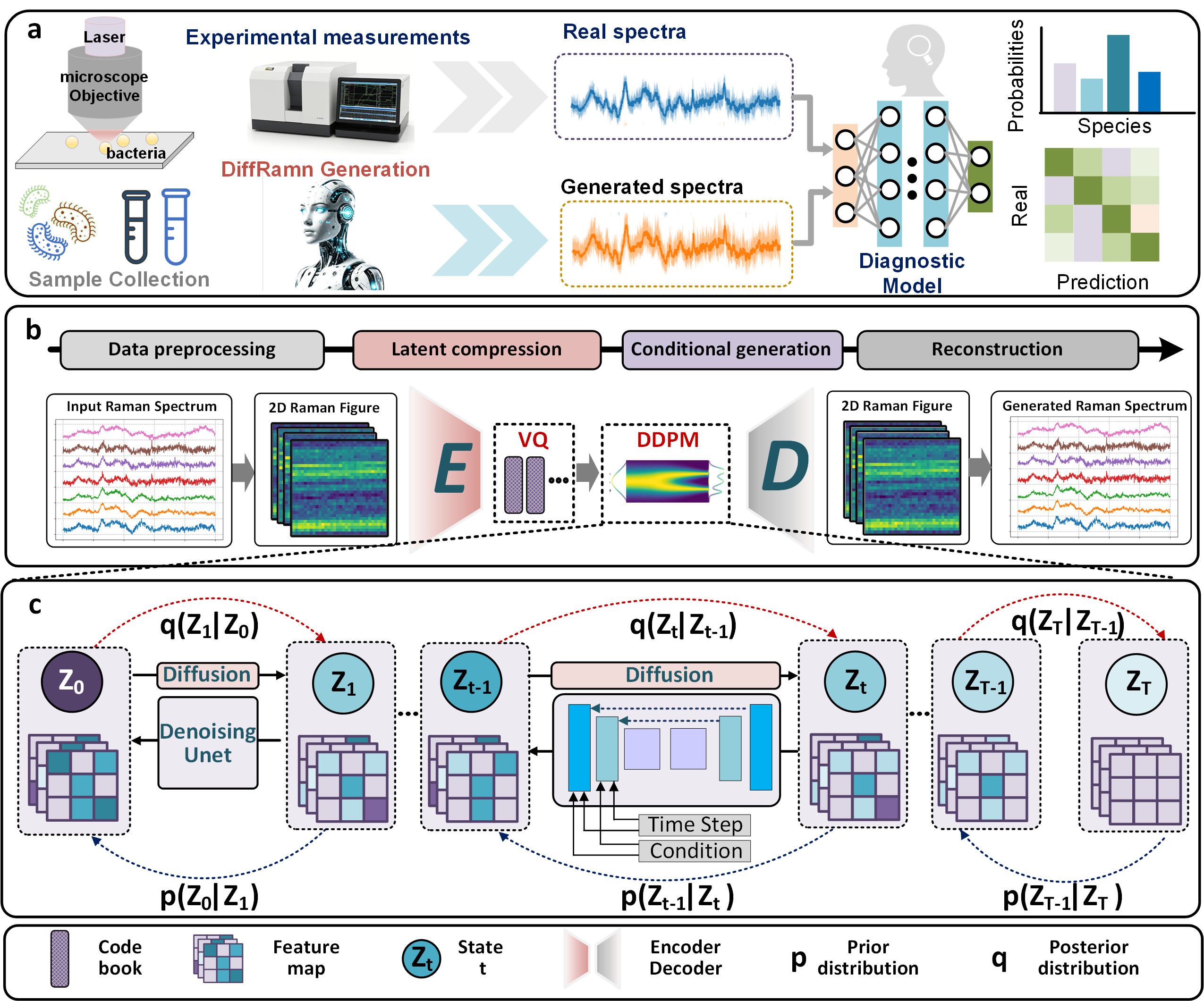}}
\caption{The proposed DiffRaman methodology: (a) DiffRaman employs a dual-data-stream approach to train diagnostic models. The first branch consists of real bacterial Raman spectroscopy data collected under actual experimental conditions, while the second branch involves synthetic bacterial Raman spectra generated by DiffRaman. The spectra from both branches are used together to train the diagnostic model, thereby achieving more robust performance.(b) The workflow of DiffRaman primarily encompasses four steps: data transformation, latent compression, conditional generation, and reconstruction. (c) Schematic illustrations of the forward noise adding and reverse denoising diffusion processes.}
\label{fig11}
\end{figure}

\section{DiffRaman Methodology}

\subsection{Overall pipeline}
To address the challenge of bacterial Raman spectroscopy diagnosis with limited data, we employed a dual-data-pathway approach as shown in Figure 1(a), which involves using both bacterial Raman spectroscopy data collected from real-world experiments and synthetic data generated by DiffRaman. The proposed DiffRaman pipeline encompasses four primary steps as illustrated in Figure 1(b) . First, the input Raman spectral data undergoes a transformation phase to convert it into a two-dimensional Raman figure. Subsequently, the VQ-VAE encoder compresses the Raman figure and extracts its latent representation, which is refined into a discrete representation using the Vector Quantization (VQ) mechanism. Thereafter, the DDPM model learns the spectral feature distribution within the compressed latent space(Figure 1(c)). Finally, the decoder reconstructs the diffusion sampling result to produce the newly generated Raman figure and Raman spectrum.

\subsection{Data Transformation}

Considering the success of DDPM in the field of two-dimensional image generation, we have developed an effective data transformation method for Raman spectra in this study as shown in Figure 2 (a). The core concept of this method is to convert the original spectral signal into a two-dimensional Raman figure. Specifically, this conversion method involves sequentially filling the original spectral signal into the pixels of the figure. To obtain a figure of size $M\times M$, an interpolated signal with a length of $M^2$ is required from the original spectral signal. Let $L(i),i=\left \{ 1,...,M^2 \right \} $, represent the values of the signal. Let $P(j,k), j=\left \{1,...,M\right \}, \left \{k=1,...,M\right \}$, denote the pixel intensity of the figure. The conversion method can be expressed by the following formula:
\begin{equation} 
P(j,k) = {\frac{{L\left({(j - 1) \times M + k} \right) - {\rm{Min}}(L)}}{{{\rm{Max}}(L) - {\rm{Min}}(L)}} \times 255}. 
\end{equation}

 The pixel values of the resulting 2D Raman figure are normalized to a range between 0 and 255, which is a standard grayscale range for images. It is recommended that $M$ be $2^n$. In this study, we use 32, implying that the length of the spectral signal is $M^2=1024$. Therefore, we interpolate the spectral signal to a length of 1024 prior to conversion. Most importantly, our proposed method involves a bidirectional mapping process, allowing us to convert Raman figure back into Raman spectra.

Some examples of transformations are shown in Figure 2(b). It is noteworthy that, unlike existing Raman image transformation methods\cite{r62}, our approach is directly effective without the need to incorporate additional expert knowledge, such as Spectral Recurrence Plot or Spectral Gramian Angular Field\cite{r62}. Furthermore, unlike these aforementioned methods, the resulting Raman figure is of low resolution, which facilitates efficient computation.

\begin{figure}[!t]
\centerline{\includegraphics[width=8.0cm]{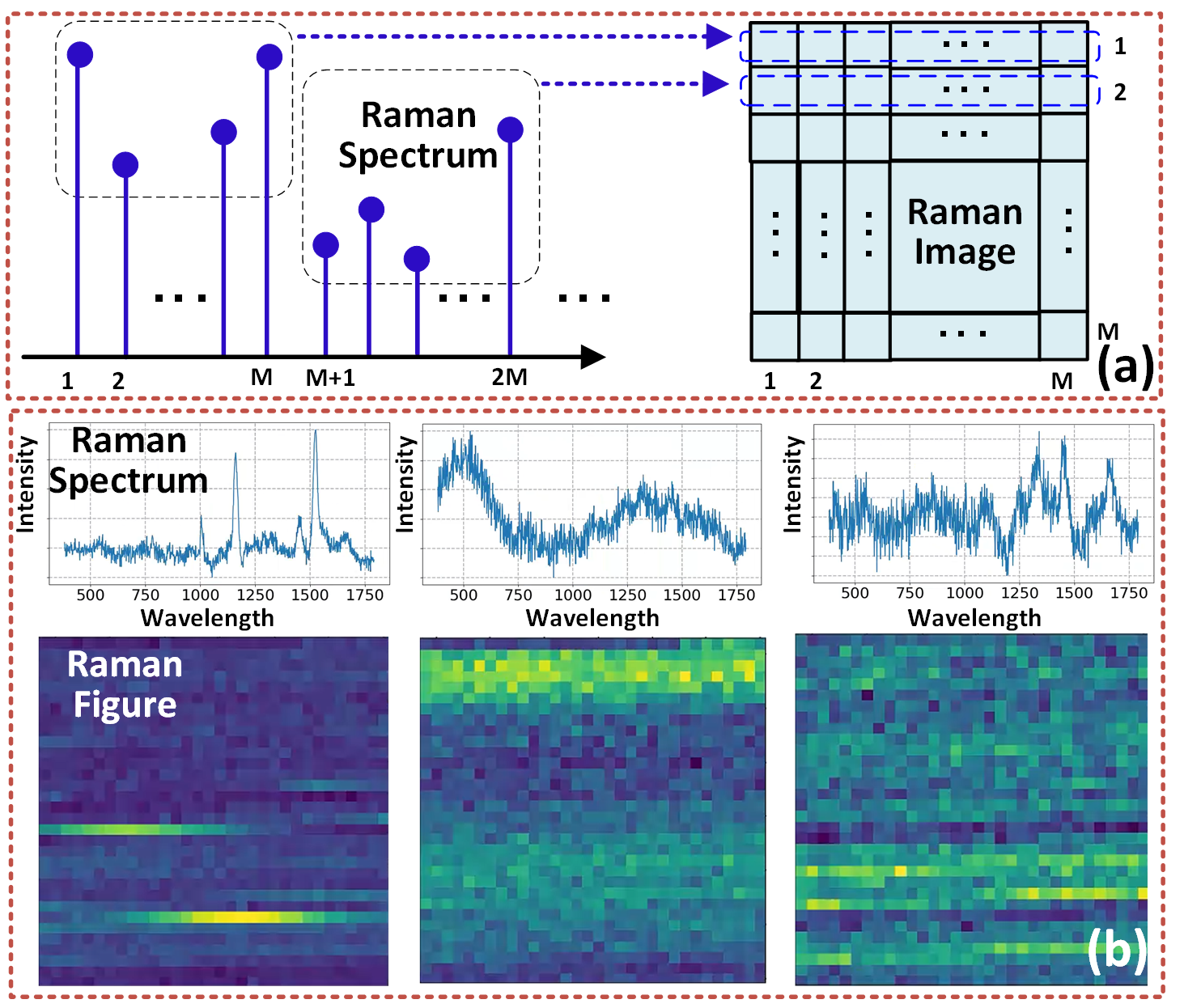}}
\caption{(a) Schematic diagram of the proposed transformation method for converting Raman spectra into two-dimensional Raman figures. (b) Examples of Raman spectra and their corresponding converted Raman figures.}
\label{fig11}
\end{figure}

\subsection{VQ-VAE for latent compression}

Given that the original DDPM is computationally expensive to apply directly in the raw data space, this study employs the dimensionality compression strategy of autoencoders to reduce the Raman figure to a latent space. Specifically, we adopt the VQ-VAE model, which employs discrete latent variable representation, proving to be more effective in numerous generative tasks\cite{r61}. Formally, the structure of the VQ-VAE consists of an encoder $E$, a decoder $D$, and a codebook $ B = \{e_n\}_{n=1}^N \in \mathbb{R}^{N \times d}$ containing a $N$ latent embedding vectors. Given an input Raman figure $P \in \mathbb{R}^{M \times M}$, the compressed continuous latent variable $z_e$ with shape of $m \times m \times d$ is obtained through the encoder as $z_e = E(P) \in \mathbb{R}^{m \times m \times d}$, and the is then quantized into $z_q$ based on its proximity to the embedding vectors in $B$:
\begin{equation} 
z_{q}(h,w)=e_{k} \in \mathbb{R}^{d}, k=\underset{n}{\arg \min }\left\|z_{e}(h,w)-e_{n}\right\|_{2}^{2}
\end{equation}
where the $(h,w)$ indicates the position in the latent feature. Finally, $z_{q}$ is input into the decoder $D$ to obtain the reconstructed Raman figure. The VQ-VAE model is trained end-to-end under several constraints. First, the decoder and encoder are optimized using reconstruction loss. Second, the vector quantization (VQ) component is optimized through dictionary learning, utilizing the $l2$ error to adjust the embedding vector 
$e_i$ towards the encoder output  $z_{e}$. Finally, a commitment loss is incorporated to ensure that the encoder adheres to the embedding and prevents its output from diverging. Consequently, the overall training objective is formulated as follows:
\begin{equation} 
\mathcal{L}_{V Q}(E,B,D)= \left\|P-D\left(z_{q}\right)\right\|_{2}^{2}+\left\|\operatorname{sg}[E(P)]-z_{q}\right\|_{2}^{2} +\zeta\left\|s g\left[z_{q}\right]-E(P)\right\|_{2}^{2}
\end{equation}
where $sg \left [  \right ] $ denotes the stop-gradient operation, and 
$\zeta $ is a hyperparameter that determines the weight of its contribution to the loss function.

\subsection{DDPM for conditional generation}

The learning process of DDPM involves two parameterized Markov chains, designed to generate samples that align with the original data distribution through a finite number of iterations. This process includes forward and reverse diffusion stages. In the forward diffusion process, Gaussian noise is incrementally added to the data, causing its distribution to converge to a specified standard Gaussian prior. Conversely, the reverse diffusion chain begins from this prior and progressively reconstructs the undisturbed data pattern.

As shown in Figure 3, the forward diffusion process in DDPM can be modeled as a Markov chain, a mathematical system that transitions between states according to a set of probabilistic rules. In this process, Gaussian noise is incrementally added to the original signal at each time step, producing potential states with a specific variance  $p$. Specifically, The $z_q$ obtained from the pre-trained VQ-VAE is denoted as the initial state $z_{q,0}$. At each time step, the process $\beta$ transitions to a new potential state through the forward noise process, adhering to the Markov principle, which can be described as:
\begin{equation} 
\begin{aligned} 
&q(z_{q,1},z_{q,2},...,z_{q,T}\mid z_{q,0})=\prod_{t=1}^{T} q(z_{q,t}\mid z_{q,t-1})\\ 
&q(z_{q,t}\mid z_{q,t-1})=\mathcal{N} \left ( z_{q,t};\sqrt{1-\beta_t}z_{q,t-1}, \beta_t\mathbf{I}  \right ) 
\end{aligned} 
\end{equation}
where $z_{q,t}$ is the latent state of the system at time step $t$. The state at time step $t$ is determined solely by the state at time step $t-1$
, denoted as $q(z_{q,t}\mid z_{q,t-1})$, and is implemented by adding Gaussian noise. This process is characterized by  $\mathcal{N} \left ( x;\mu , \sigma  \right )$, where  $\mu $  is the mean and $\sigma $ is the variance that produces $x$. Employing the reparameterization trick, the state $z_{q,t}$ at a specific time step $t$ can be directly derived from $z_{q,0}$:
\begin{equation} 
q(z_{q,t}\mid z_{q,0})=\sqrt{\bar{\alpha}_t  } z_{q,0}+\sqrt{1-\bar{\alpha}_t} \varepsilon ,\varepsilon \in \mathcal{N}\left ( 0,\mathbf{I} \right )  
\end{equation}
where $\alpha_t=1-\beta_t$ and $\bar{\alpha}_t =\prod_{s=1}^{t} \alpha_s$. It can be observed that when $\bar{\alpha}_t $ approaches zero, $q(z_{q,t}\mid z_{q,0})$ will converge to the Gaussian prior distribution $\mathcal{N}\left ( 0,\mathbf{I} \right )$.

\begin{figure}[!t]
\centerline{\includegraphics[width=8.0cm]{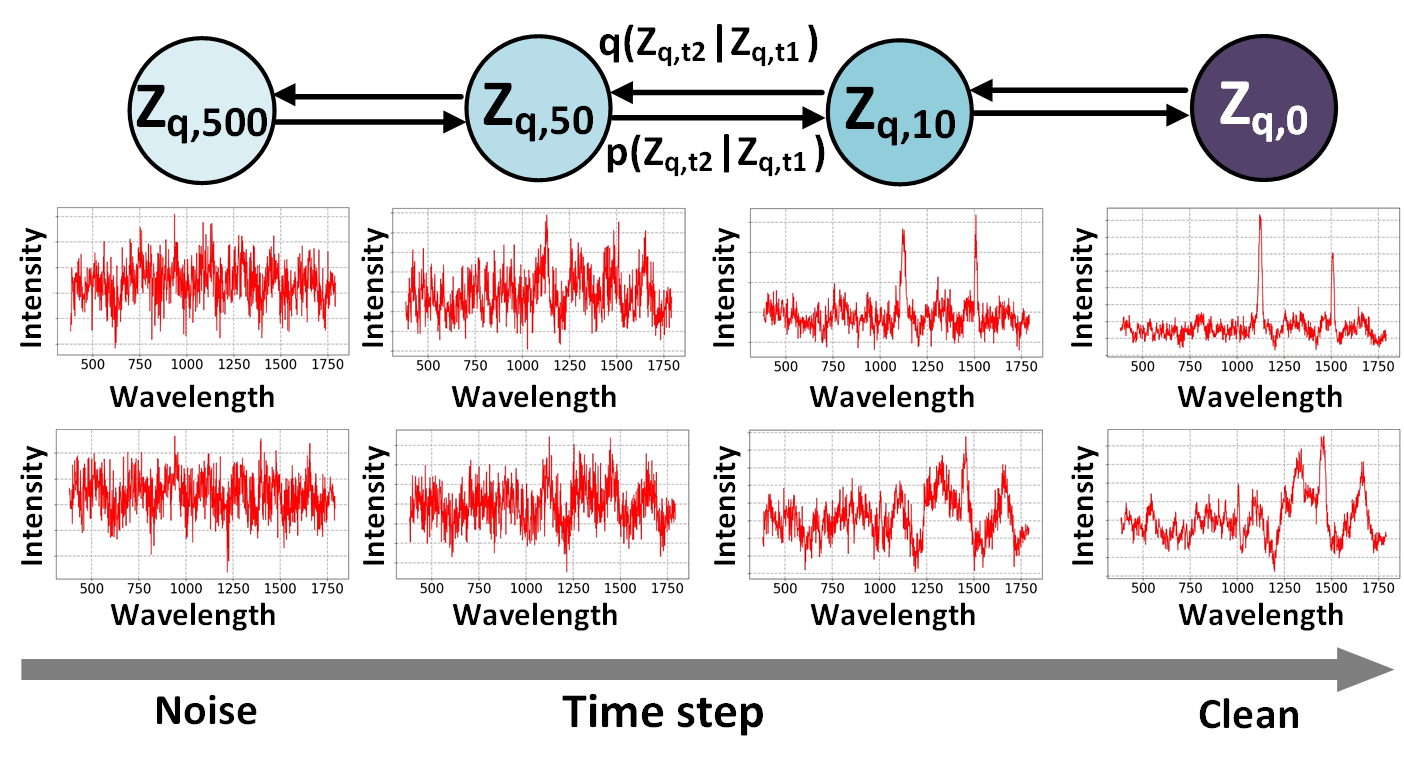}}
\caption{Schematic illustration of forward noise addition and reverse denoising for Raman spectroscopy based on Markov chain.}
\label{fig11}
\end{figure}

The objective of the reverse diffusion process in DDPM is to learn the distribution $p(z_{q,t-1}\mid z_{q,t})$, which is equivalent to a denoising process as shown in Figure 3. Additionally, to achieve controllable generation of different types of Raman spectra, we further employ the conditional distribution $p(z_{q,t-1}\mid z_{q,t}, y)$ for modeling, given a spectrum type label $y$. In this context, $z_{q,T}$ can be randomly sampled from $\mathcal{N}\left ( 0,\mathbf{I} \right )$ to generate $p(z_{q,0}\mid z_{q,T}, y)$ as the synthetic representation. For this purpose, a neural network parameterized by $\theta$ is applied to approximate this process:
\begin{equation} 
\begin{aligned} 
&P_{\theta }(z_{q,0,...,T}, y)=P( z_{q,T})\prod_{t=1}^{T} P_{\theta }( z_{q,t-1}\mid  z_{q,t}, y)\\ 
&P_{\theta }( z_{q,t-1}\mid  z_{q,t}, y) = \mathcal{N}\left ( z_{q,t-1}; \mu_\theta ( z_{q,t},t, y), \Sigma_{\theta }( z_{q,t},t, y)\right ) 
\end{aligned} 
\end{equation}
where the $\mu_\theta ( z_{q,t},t, y)$ and $\Sigma_{\theta }( z_{q,t},t, y) ) $ represent the learnable mean and variance functions at the reversed step $t$, conditioned on the $y$. The optimization of $\theta$ is achieved by maximizing the variational lower bound of the negative log-likelihood of the data distribution $p(z_{q,0})$:
\begin{equation} 
\begin{array}{l}
\max _{\theta} \mathbb{E}_{q\left(z_{q,0}\right)}\left[\log p_{\theta}\left(z_{q,0},y\right)\right] \\
\leq \max _{\theta} \mathbb{E}_{q\left(z_{q,0}, \ldots, z_{q,T}\right)}\left[\log p_{\theta}\left(z_{q,0: T},y\right)-\log q\left(z_{q,1: T} \mid z_{q,0}\right)\right] \\
=\max _{\theta} \mathbb{E}_{q\left(z_{q,0}, \ldots, z_{q,T}\right)}\left[-\log p\left(z_{q,T}\right)-\sum_{t \geq 1} \log \frac{p_{\theta}\left(z_{q,t-1} \mid z_{q,t},y\right)}{q\left(z_{q,t} \mid z_{q,t-1}\right)}\right]
\end{array}
\end{equation}

As described in \cite{r63}, the above formula can be transformed into the following loss function for optimization:
\begin{equation} 
\mathcal{L}_{DDPM}\left(\varepsilon_{\theta}\right)=\mathbb{E}_{z_{q,0}, \varepsilon, t}\left[\left\|\varepsilon_{t}-\varepsilon_{\theta}\left(z_{q,t}, t,y\right)\right\|_{2}^{2}\right]
\end{equation}
where the $\varepsilon_{\theta}\left(z_{q,t}, t,y\right)$ is a trainable U-Net neural network\cite{r64} conditioned on the time step $t$ and class label $y$ that can predict the noise $\varepsilon_t$.
\subsection{Training and Inference}

The training and optimization process of the proposed DiffRaman framework can be divided into two distinct stages. In the first stage, the VQ-VAE is trained using Equation (3). Subsequently, in the second stage, the trained VQ-VAE is utilized to effectively compress the Raman figure into a low-dimensional latent space, and the DDPM is optimized using Equation (8).

\begin{algorithm}[t]
\caption{DiffRaman training algorithm}

\SetKwInOut{Input}{input}
\SetKwInOut{KwOut}{Hyper-parameters}
\KwIn{Spectral data $L$; Spectral type label $y$}
 \KwOut{Reconstruction epoch $E_1$, Diffusion epoch $E_2$, Diffusion step $T$}
\KwResult{Encoder $E$, codebook $ B$, Decoder $D$, and U-Net noise predictor $\varepsilon_{\theta}$}
  
 \textbf{$\#$ Stage I: VQ-VAE Training}\;
\For{$epoch=1,...,{E_1}$}
{
Calculate loss: $\mathcal{L}_{VQ}(E,B,D)\gets \mathrm{Eq.(3)}$\;

Update the Encoder $E$, codebook $ B$, Decoder $D$ by backpropagation;

}

 \textbf{$\#$ Stage II: DDPM Training}\;
\For{$epoch=1,...,{E_2}$}
{
Sample $t \sim \mathrm{Uniform} (1,...,T)$\;
Calculate loss: $\mathcal{L}_{DDPM}(\varepsilon_{\theta})\gets \mathrm{Eq.(8)}$\;

Update the U-Net noise predictor $\varepsilon_{\theta}$ by backpropagation;

}
return $E$,$ B$, $D$, $\varepsilon_{\theta}$
\end{algorithm}

\begin{algorithm}[t]
\caption{DiffRaman sampling algorithm}

\SetKwInOut{Input}{input}
\SetKwInOut{KwOut}{Hyper-parameters}
\KwIn{Spectral type label $y$}
 \KwOut{Diffusion step $T$}
\KwResult{Generated Spectral data}

Sample $z_{q,T}\sim\mathcal{N}(0, \mathrm{I})$\;
\For{$t=T,...,1$}
{
Calculate the $z_{q,t-1}$ according to Eq. (9)

}
return $D(z_{q,0})$
\end{algorithm}

After completing the optimization process, the trained system can be employed for spectrum generation. Initially, a sample is randomly drawn from Gaussian distribution $\mathcal{N}\left ( 0,\mathbf{I} \right )$ to serve as $z_{q,T}$ for step $T$. This data is then processed through the trained model to obtain the predicted noise $\varepsilon_{\theta}\left(z_{q,t}, t,y\right)$ conditioned on the time step $t$ and class label $y$. To derive noise-free data, the process iteratively continues backward to time step $0$, where sampling is conducted from $t$ to $t-1$ according to the following formula:
\begin{equation} 
z_{q,t-1}=\frac{1}{\sqrt{\alpha_{t}}}\left(z_{q,t}-\frac{\beta_{t}}{\sqrt{1-\bar{\alpha}_{t}}} \varepsilon_{\theta}\left(z_{q,t}, t,y\right)\right)+\sigma_{t} z, z \sim \mathcal{N}(0, \mathrm{I})
\end{equation}
where $\sigma_{t}= \sqrt{\beta_t}$. Subsequently, $z_{q,0}$ is fed into the decoder $D$ to obtain the reconstructed Raman figure and spectra. The training and sampling process can be described more compactly by Algorithms 1 and 2.

\section{Results and Discussion}

\subsection{Material}
This study utilized two well-recognized bacterial spectral datasets: Bacterial ID\cite{r1} and the Bacterial Strains Dataset\cite{r12} for the purpose of experimental validation. Detailed information concerning these datasets is as follows.

\begin{figure}[!t]
\centerline{\includegraphics[width=15.0cm]{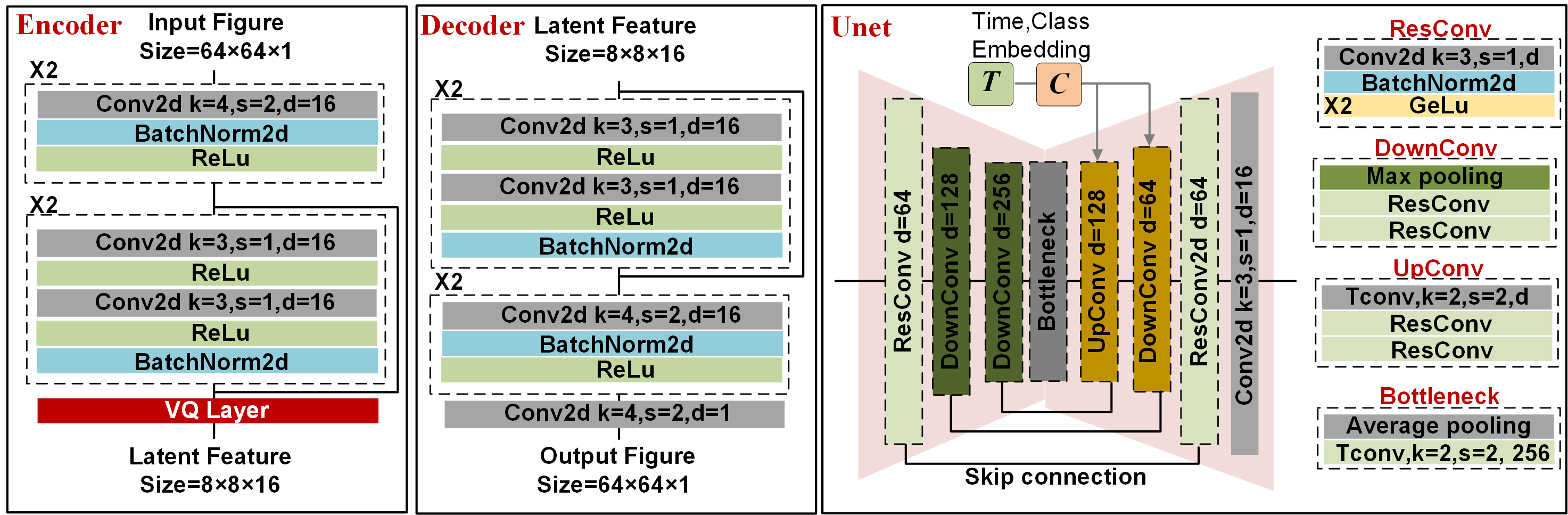}}
\caption{The specific network architectures of the encoder, decoder, and UNet components used in DiffRaman.}
\label{fig11}
\end{figure}

\subsubsection{Bacteria-ID dataset}
We first utilized the large-scale Bacteria-ID dataset\cite{r1}. This dataset comprises Raman spectral data from 30 different bacteria and yeasts, encompassing most infections commonly found in intensive care units worldwide. The wave number range of the spectra spans from 381.98 to 1792.4 cm$^{-1}$. The dataset is divided into three subsets: a reference dataset, a fine-tuning dataset, and a test dataset. The reference dataset includes 2000 spectral data for each of the 30 bacterial types, totaling 60,000 spectral data. In addition to the reference dataset, the fine-tuning and testing datasets include 3,000 spectral samples covering 30 bacterial classes.  In the existing literature, researchers typically pre-train the model utilizing the large-scale reference dataset, followed by fine-tuning with the fine-tuning dataset. Nevertheless, to assess the model's diagnostic performance under limited data conditions, we have chosen to exclude the reference dataset and only train the model on a restricted subset of the fine-tuning dataset. For additional information regarding this dataset, please refer to \cite{r1}.

\subsubsection{Bacterial Strains dataset}

The bacterial strain dataset represents another comprehensive collection of bacterial Raman spectroscopy data, encompassing 9 strains across 7 species. The authors employed five distinct laser acquisition durations, specifically 0.01, 0.1, 1, 10, and 15 seconds, to measure approximately 250 cells for each strain at each acquisition interval. This procedure yielded around 2,250 spectral samples, culminating in a total of 11,141 single-cell Raman spectra across the five measurement conditions. According to the original paper, acquisition times of 0.01 and 0.1 seconds result in a signal-to-noise ratio (SNR) that is insufficient for reliable diagnosis. Conversely, acquisition times of 10 or 15 seconds result in an excessively high SNR, rendering the diagnostic task overly simplistic and failing to adequately represent measurement noise interference typical of real experimental conditions. Taking these factors into account, we opted for the dataset with a measurement duration of 1 second, and randomly selected 60$\%$ of the data for the training set and 40$\%$ for the test set. Similarly, to simulate the scenario of limited sample sizes, we only train the model on a selected subset of the training set. 
For further details regarding the dataset, please refer to \cite{r51}.

% The bacterial strain dataset is another large-scale bacterial Raman spectroscopy dataset, which includes 9 strains of 7 species. The authors used five different laser acquisition times, set to 0.01, 0.1, 1, 10 or 15 seconds, and measured about 250 cells for each strain at each acquisition time, acquiring about 2250 spectral samples, and a total of 11,141 single-cell Raman spectra were obtained under the five measurement conditions. As described in the original paper, when the measurement time is 0.01 and 0.1, the SNR of the acquired signal is very low and is not suitable for effective diagnosis. When the measurement time is 10 or 15 seconds, the SNR of the signal is very high, making the diagnostic task too simple and unable to reflect the measurement noise interference in the real experimental environment. Considering all factors, we chose the dataset with a measurement time of 1 second, and randomly selected 60$\%$ of the data as the training set and 40$\%$ of the data as the test set.

\subsection{Experimental implementation details}

The detailed network architecture of the encoder $E$, U-Net noise predictor $\varepsilon_{\theta}$, and decoder $D$ components within the DiffRaman framework is illustrated in Figure 4. "Conv2d" and "Tconv2d" denote the two-dimensional convolutional layer and its transposed counterpart, respectively. Here, $k$ signifies the size of the convolutional kernel, $s$ represents the stride utilized in the convolution process, and $n$ denotes the number of convolutional kernels employed in each layer.The total count of latent embedding vectors encompassed within Codebook $B$ amounts to 1024. In training the parameters of DiffRaman, we employed the Adam optimization algorithm with a learning rate of $\eta = 1 \times 10^{-3}$. The maximum number of training iterations for the two stages, $E_1$ and $E_2$, were both set to 600 and 1000. The batch size $BS$ used during training was 32. Additionally, the diffusion steps $T$, were set to 500, with the variance $\beta$ increasing linearly from $1 \times 10^{-4}$ to $0.02$. The hyper-parameters of training are shown in Table 1. All experiments were implemented using PyTorch 1.12 and executed on a system equipped with Intel(R) Xeon(R) Gold 5220 CPUs running at 2.20 GHz, along with an NVIDIA 4090 GPU featuring 24 GB of memory.

In the experiments, we considered three data generation techniques as comparative methods, including:  (1). Data augmentation(DA), which applied various perturbation techniques to the original spectra, such as random Gaussian blur, random Gaussian noise, and random scaling. (2). Conditional Variational Autoencoder (cVAE)\cite{r53}. (3). Conditional Deep Convolutional Generative Adversarial Networks (cDCGAN)\cite{r68}. In order to simulate diagnostic scenarios with a limited number of samples, we selected training subsets from two bacterial Raman datasets. For each dataset, we chose two data scales, one containing 300 real samples and the other 600 real samples. To ensure class balance, uniform sampling was performed across all bacterial categories. The above four data generation methods—DA, cVAE, cDCGAN, and DiffRaman—were employed to generate new synthetic data based on the selected training subsets. Without further specification, we have chosen the recently advanced SANet\cite{r16} as the diagnostic model. To ensure a fair comparison, all the considered generation methods are trained on the same real samples. Additionally, the diagnostic models derived from different generation methods are all trained using an early stopping strategy with a patience of 10 epochs on the test loss for fair comparison.

\begin{table}
\caption{{Training hyperparameters of DiffRaman}}
\resizebox{80mm}{!}{

\begin{tabular}{cll}
\Xhline{1.5pt}
Notation & Definition   & Value                                         \\ 
\hline
$\eta$     & Learning rate     & $1 \times 10^{-3}$                     \\
$E_1$        &  Reconstruction epoch     & 600                   \\
$E_2$        &  Diffusion epoch     & 1000                      \\
$T$        & Diffusion step & 500   \\
$\beta$        & Variance schedule        &  Linear($1 \times 10^{-4}$, 0.02)                   \\
$BS$        & Batch size             & 32     \\                             
\Xhline{1.5pt}
\end{tabular}

}
\label{table1}
\end{table}

\subsection{Results}

\subsubsection{Spectrum Generation Evaluation}

The quality of generated samples is a key factor. This study conducted a comprehensive evaluation from multiple perspectives. Firstly, a visual comparison of the Raman spectra and figures between the generated samples and the real samples was performed. Secondly, to more accurately assess the quality of the generated samples, we employed quantitative evaluation metrics to measure the similarity between the generated samples and the real samples. In this experiment, we utilized 600 real samples and generated an additional 600 synthetic samples using the data synthesis methods.

In Figure 5, we present the average Raman images of eight classes from the two datasets, both from real experiments and generated using various methods. The findings suggest that the average Raman images generated by DA and cVAE are very similar to the average Raman images obtained from real experiments. In contrast, the average Raman images produced by cDCGAN show a higher level of noise. The spectra generated by DiffRaman are generally analogous to the real average Raman images, yet they do not completely overlap. Furthermore, in Figure 6, we perform an inverse transformation of the generated Raman images back into Raman spectra and give the distribution for each category. Based on these two sets of outcomes, we can infer the following conclusions:

(1). The samples generated by DA exhibit a high degree of overlap with the overall distribution of real samples and demonstrate diversity. However, this technique relies on primary signal manipulation methods such as simple scaling, Gaussian noise addition, and filtering, which do not accurately reflect the real data distribution. (2). Samples generated by cVAE tend to replicate the average pattern of real samples and lack the capability to generate diverse samples. (3). The generation process of cDCGAN encounters mode collapse due to the limited number of real samples. This not only results in generated samples deviating from the distribution of real samples and being contaminated with noise, but also the diversity of patterns in the generated data is very limited. (4). Conversely, DiffRaman generates samples with a high degree of diversity while ensuring that the overall distribution remains unbiased.

\begin{figure}[!t]
\centerline{\includegraphics[width=15.0cm]{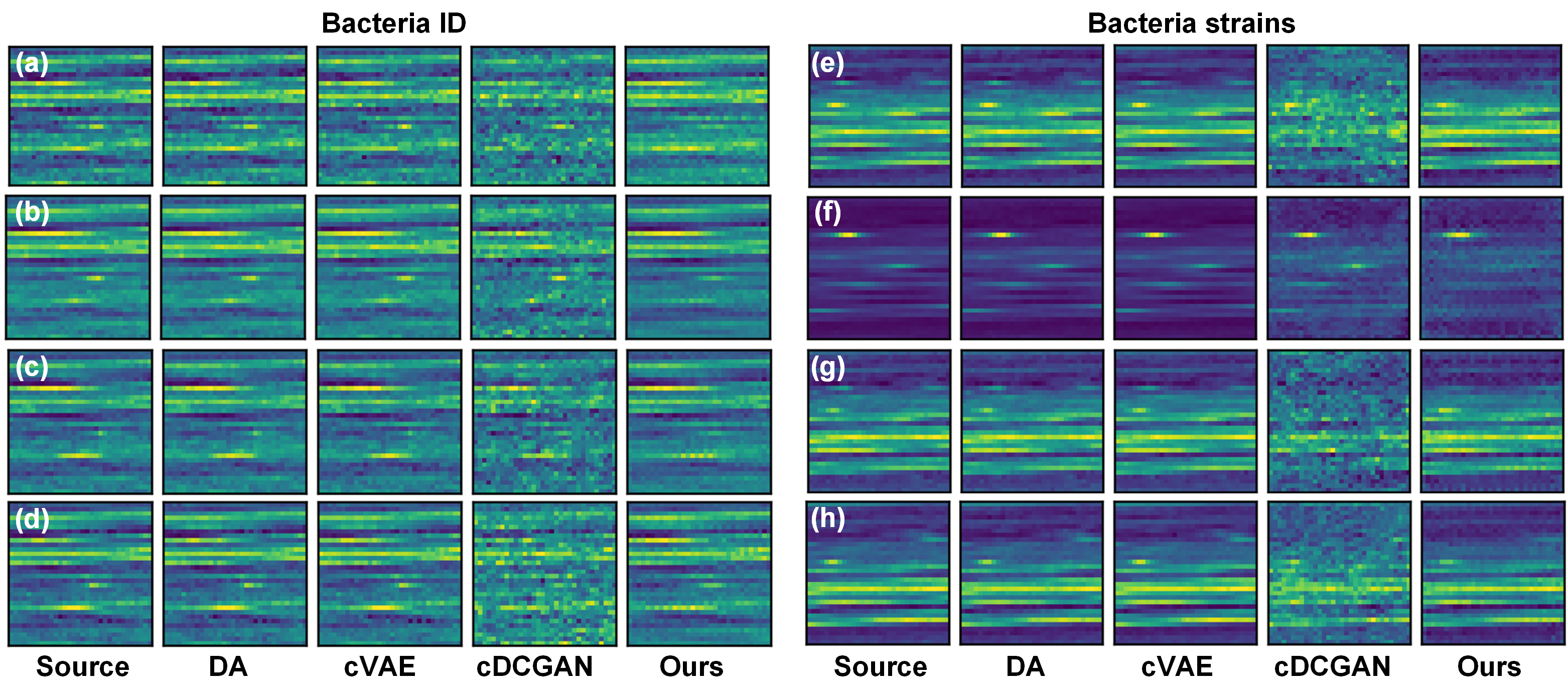}}
\caption{A comparative example of real and generated Raman figures. (a)-(d): Isolate-averaged Raman figure for E. faecalis 2, MSSA 3, S. lugdunensis, and Group C Strep. in Bacteria ID dataset. (e)-(h): Species-averaged Raman figure for S. aureus,  P. aeruginosa, S. epidermidis, and E. faecalis in Bacteria stains dataset.}
\label{fig12}
\end{figure}

 % In contrast, the spectra generated by CGAN display significant distributional divergence from the original data, suggesting that CGAN fails to accurately capture the true distribution of the original spectral data. As a result, the usability of CGAN-generated data in practical applications is considerably limited.

\begin{figure}[!t]
\centerline{\includegraphics[width=15.0cm]{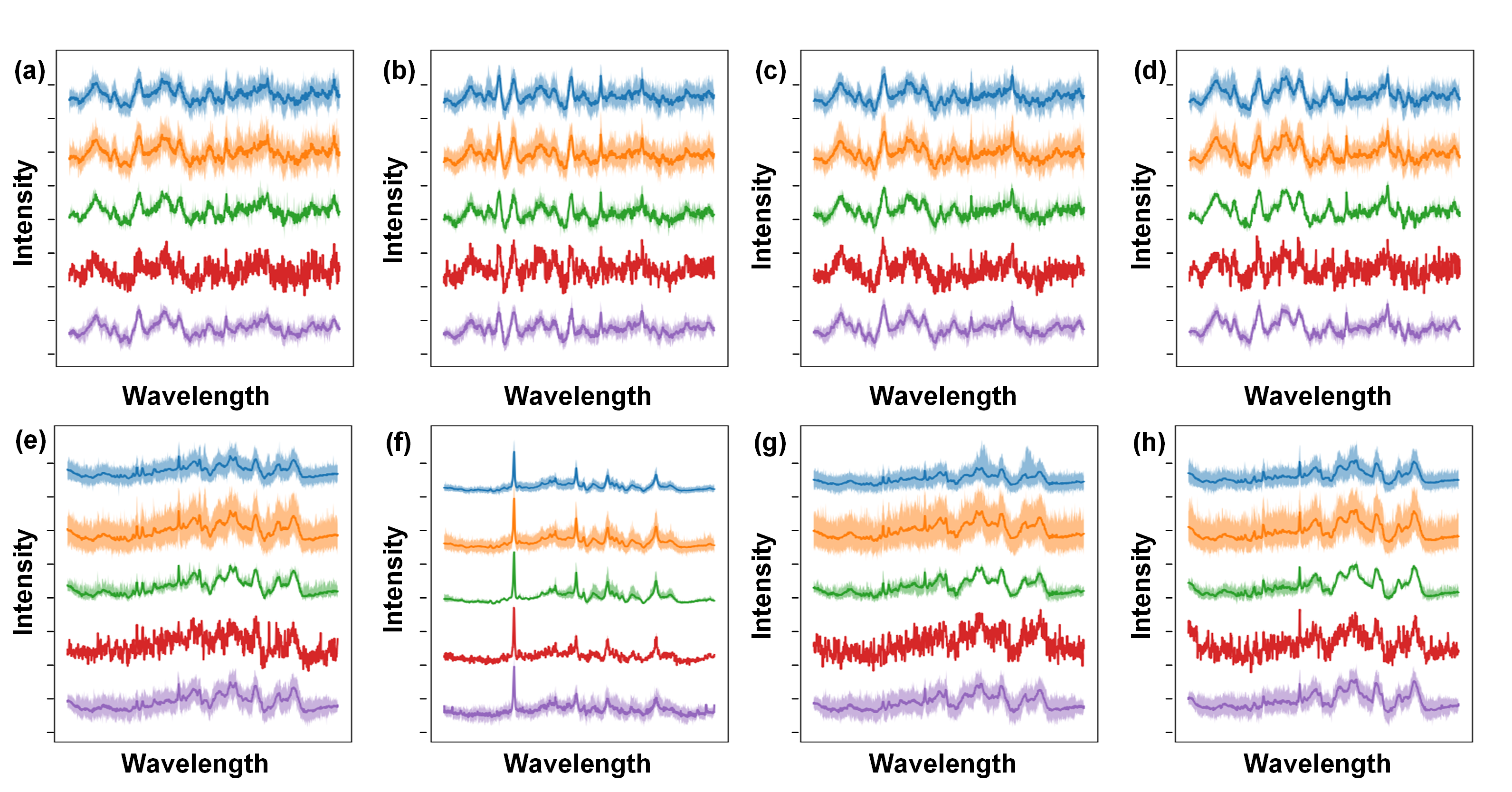}}
\caption{A comparative example of real and generated  Raman spectra. (a)-(d): Raman figure for E. faecalis 2, MSSA 3, S. lugdunensis, and Group C Strep. in Bacteria ID dataset. (e)-(h): Raman spectra for S. aureus,  P. aeruginosa, S. epidermidis, and E. faecalis in Bacteria stains dataset. From top to bottom: real spectra, and synthesized spectra generated by DA, cVAE, cDCGAN, and DiffRaman. The lines represent the average spectra, and the shading indicates the variation range within the class.}
\label{fig11}
\end{figure}

% To assess the data generation performance of DiffRaman, we validated it on the large-scale bacterial dataset. The relevant results are illustrated in Figure 3. In Figure 3(a), we first demonstrated the effectiveness of the proposed data pre-processing method, which transforms the one-dimensional Raman spectrum into a two-dimensional Raman figure. From the converted figures, it is evident that different bacterial Raman spectra exhibit distinct image textures, offering a more intuitive observation method than a one-dimensional spectrum. To intuitively illustrate the Markov inverse diffusion denoising generation process of DDPM, we present a relevant example in Figure 3b. It can be observed that the generated spectra, obtained using the system's latent state at different time steps 
% $t$, vary significantly. When $t$ is large, such as $t=500$ and $t=100$, the semantic pattern information of the generated spectrum is weak, and the signal is largely obscured by noise. As 
% $t$ decreases, such as $t=10$, the generated spectrum begins to exhibit stronger semantic information, albeit with some residual noise interference. At $t=0$, the generated spectrum displays a clear and distinct spectral pattern.

To quantitatively assess the quality of the generated signals, we employ similarity evaluation metrics. Specifically, to evaluate the similarity between individual samples, we utilize the Cosine Similarity (COS), which calculates the average cosine similarity between each pair of generated and real samples. To assess the overall similarity of synthetic samples, we use distribution similarity metrics, including Maximum Mean Discrepancy (MMD), Jensen-Shannon Divergence (JS), and Wasserstein Distance (WD). It is noteworthy that these three distribution metrics are all conducted under the assumption of multivariate Gaussian distribution. The higher the COS value, the greater the similarity between the signals; conversely, the lower the values of MMD, JS, and WD, the higher the similarity between the distributions.

Table 2 presents the comparative results of different methods on two datasets. The symbol "$\uparrow$" in the table indicates that higher values correspond to greater similarity, while "$\downarrow$" indicates that lower values correspond to greater similarity. It can be observed that out of the 8 comparisons across 4 metrics on two datasets, the proposed DiffRaman achieved 7 optimal or sub-optimal results, cVAE achieved 5, DA achieved 4, and cDCGAN performed the worst. This suggests that the spectral data generated by DiffRaman closely matches the original spectra in both sample similarity and distribution similarity. In contrast, DA-generated samples only exhibit similarity in overall distribution but have lower sample similarity, while cVAE-generated spectra have high sample similarity but low overall distribution similarity. cDCGAN performs poorly in both aspects. These results align with the aforementioned qualitative analysis, indicating that DA, cVAE, and cDCGAN generate lower quality data, reducing their usability for subsequent classification tasks.

\begin{table}
\caption{Quantitative assessment results of Raman spectroscopy generation quality}
\centering

\scriptsize

\begin{tabular}{lcccccccc}
\toprule
Method & \multicolumn{4}{c}{Bacteria ID} & \multicolumn{4}{c}{Bacteria strains} \\
\cmidrule(lr){2-5} \cmidrule(lr){6-9}
     & COS($\uparrow$) & MMD($\downarrow$) & JSD($\downarrow$) & FD($\downarrow$) & COS($\uparrow$) & MMD($\downarrow$) & JSD($\downarrow$) & FD($\downarrow$) \\
\midrule
DA & 0.676 & \textbf{0.078} & \textbf{0.469} & 21.022 & 0.636 & \textbf{0.398} & \underline{4.520} & 59.547 \\
cVAE & \textbf{0.789} & 0.405 & \underline{0.722} & \textbf{11.323} & \underline{0.686} & 0.861 & 4.824 & \underline{54.904} \\
cDCGAN & 0.699 & 0.686 & 2.030 & 17.434 & 0.632 & 1.195 & 5.273 & 62.315 \\
Ours & \underline{0.744} & \underline{0.152} & 1.046 & \underline{13.756} & \textbf{0.713} &  \underline{0.403} & \textbf{4.455} & \textbf{51.351} \\
\bottomrule

\end{tabular}

\begin{tablenotes}
       \footnotesize
        \item[1]The most favorable outcomes are highlighted in bold, while the second-best results are indicated with underlining.
\end{tablenotes}

\end{table}

\subsubsection{Spectrum Identification Evaluation}

In this section, we conducted recognition experiments to assess the impact of spectra generated by different methods on the diagnostic model performance under limited sample conditions. Specifically, we employed two distinct processes. The first is the basic method, where the diagnostic model is trained solely on source data (Source). The second process involves training the diagnostic model with both source data and synthetic data, where the synthetic data is generated through the four methods mentioned above: DA,cVAE,cDCGAN, and the proposed DiffRaman. To avoid accidental results when sampling a small number of instances and to ensure the reliability of the findings, the experimental results presented are the mean of five independent trials.

\begin{figure}[!t]
\centerline{\includegraphics[width=15.0cm]{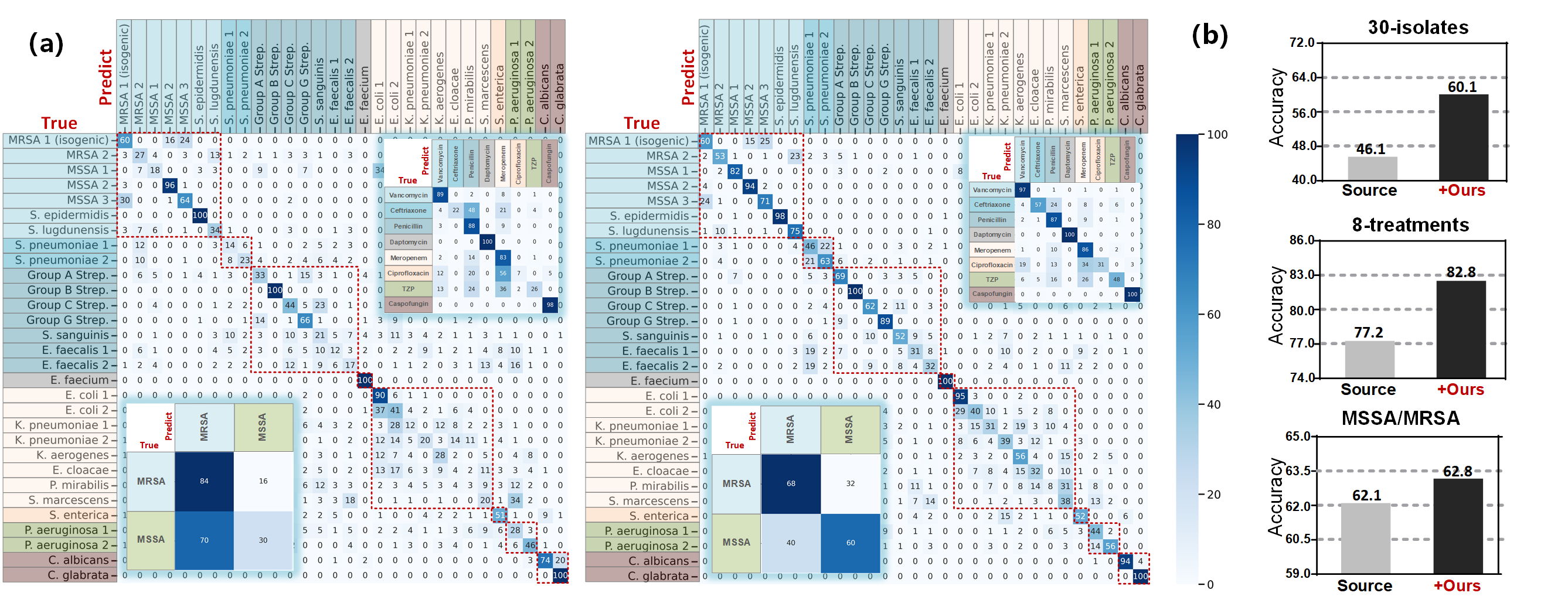}}
\caption{On the Bacterial ID dataset, with only 300 real bacterial Raman spectra used, the enhancement of diagnostic results using 600 synthetic spectra generated by the proposed DiffRaman. (a) Confusion matrices for three diagnostic tasks using only real data and in conjunction with synthetic data. (b) Improvement in accuracy for three diagnostic tasks.}
\label{fig11}
\end{figure}

Firstly, we reported the diagnostic performance on the bacterial ID dataset. This dataset encompasses three diagnostic tasks: (1) the identification of 30 isolates, (2) the classification of the 30 bacterial isolates into 8 empirical treatment groups, and (3) the categorization of Staphylococcus aureus in the dataset into two binary groups based on methicillin susceptibility: methicillin-resistant Staphylococcus aureus (MRSA) and methicillin-susceptible Staphylococcus aureus (MSSA). When using 300 real samples, we generated an additional 600 samples with our DiffRaman method. The improvement in diagnostic results using synthetic samples across the three diagnostic tasks is shown in Figure 7. 
As shown in the confusion matrix of Figure 7(a), when only 300 real experimental samples are used (10 samples per isolate), the diagnostic model performs poorly on any of the three tasks.In the identification task of 30 isolates, the accuracy of some categories such as Streptococcus pneumoniae type 2, MSSA type 1, and Klebsiella aerogenes was very low, not exceeding 30$\%$, and the overall accuracy was only 46.1$\%$. In the diagnosis task of 8 treatment groups, the accuracy of ciprofloxacin, ceftriaxone, and piperacillin-tazobactam (TZP) did not exceed 30$\%$. In the MSSA/MRSA binary classification task, there was a serious problem of misclassifying MSSA as MRSA. When the diagnostic model was trained using 600 additional samples generated by DiffRaman combined with the original real experimental data, the diagnostic accuracy of the model was significantly improved, and the diagnostic performance of some categories was significantly enhanced. For example, in the identification task of 30 isolates, the diagnostic accuracy of MSSA 1 was increased from 18$\%$ to 82$\%$. In the task of classifying eight treatment groups, the diagnostic accuracy of ceftriaxone was improved from 22$\%$ to 57$\%$. As shown in Figure 7(b), the average overall accuracy of the three tasks for five trials was improved by +14.2$\%$, +3.9$\%$, and +0.7$\%$, respectively, demonstrating its effectiveness.

\begin{table}[h]
\caption{Accuracy of different spectral generation methods on the three diagnostic tasks of the Bacterial ID dataset}
\centering
\scriptsize
\begin{tabular}{ccccccccccc}
\toprule
Task & \multicolumn{5}{c}{300 Real +600 Generation Samples} & \multicolumn{5}{c}{600 Real +1200 Generation Samples} \\ \cmidrule(lr){2-6} \cmidrule(lr){7-11}
 & Source & DA & cVAE & cDCGAN & Ours & Source & DA & cVAE & cDCGAN & Ours \\\midrule
30-isolates & 0.461 & \underline{0.466} &  0.445 &  0.452 & \textbf{0.603} 
&  0.654 & \underline{0.667} &  0.642 &  0.635 & \textbf{0.756} \\
8-treatments & 0.752 & \underline{0.789} &  0.767 &  0.771 
& \textbf{0.828} &  0.874 & 0.885 &  \underline{0.892} &  0.878 & \textbf{0.932} \\
MSSA/MRSA & \underline{0.621} & 0.590 &  0.615 &  0.604 & \textbf{0.628} 
& 0.636 & 0.621 &  0.634 & 0.646 & \textbf{0.664} \\ \hline
Average & 0.611 &  \underline{0.615} & 0.609 & 0.609 &  \textbf{0.667} 
& 0.721 & \underline{0.724} & 0.722 & 0.719 & \textbf{0.784} \\ 
\bottomrule
\end{tabular}
\label{tab:training_data_comparison}
\end{table}

Furthermore, we compared the impact of data generated by other models on diagnostic performance. We employed two experimental setups: the first used 300 real samples and 600 generated samples for training the diagnostic model, while the second used 600 real samples and 1200 generated samples for training. The results are reported in Table 3. The conclusions that can be drawn are: (1) Our method outperforms other data generation methods in enhancing the performance of the diagnostic model across the three tasks under both setups, especially in the identification task of 30 isolates, where our method surpasses the next best method (DA) by accuracy rates of 13.7 $\%$ and 8.9$\%$.(2) Not all methods can enhance diagnostic accuracy. In the identification task of 30 isolates, the data generated by cVAE and cDCGAN have a negative impact on the diagnostic model. This is due to the low quality of the generated data, which makes it unsuitable for application in downstream identification tasks.

\begin{figure}[!t]
\centerline{\includegraphics[width=15.0cm]{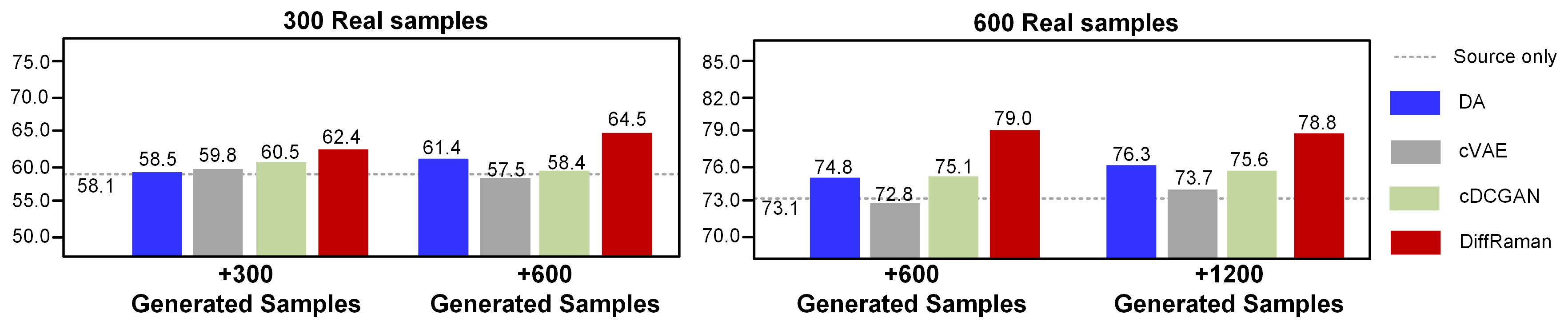}}
\caption{Diagnostic accuracy results of different spectral generation methods on the Bacterial Strains dataset.}
\label{fig11}
\end{figure}

\begin{figure}[!t]
\centerline{\includegraphics[width=15.0cm]{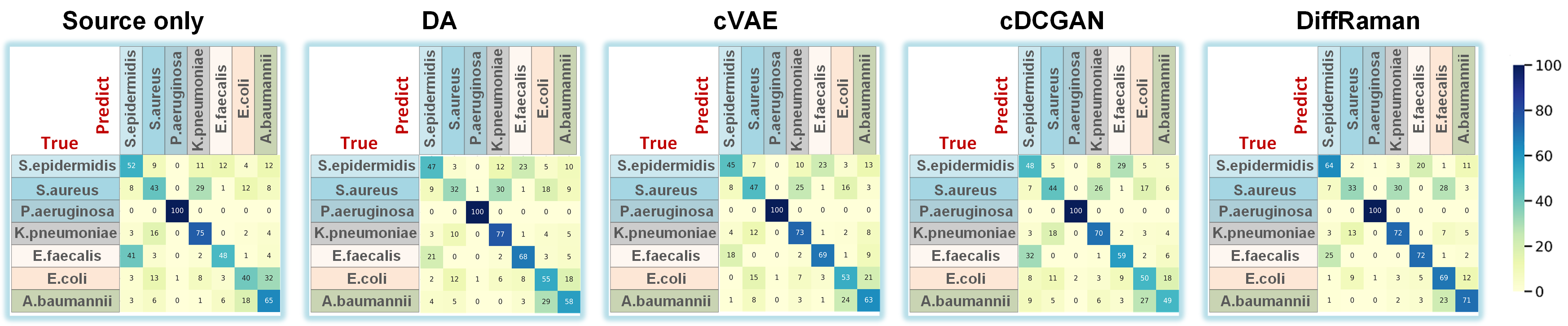}}
\caption{Confusion matrices of different spectral generation methods on the bacterial strain dataset.}
\label{fig11}
\end{figure}

\begin{figure}[!t]
\centerline{\includegraphics[width=15.0cm]{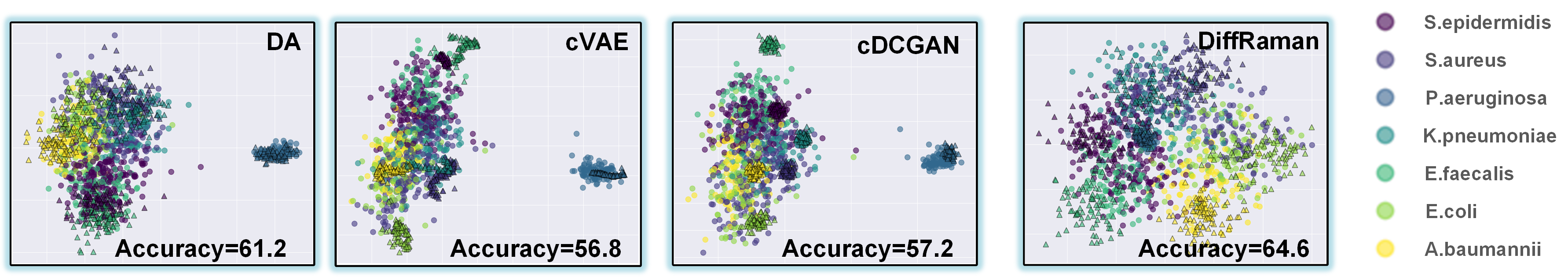}}
\caption{PCA scatter plots of different generation methods on the Bacterial Strains dataset. Triangles represent synthetic spectral features obtained by generation methods, while circles represent real experimental spectra in the test set.}
\label{fig11}
\end{figure}

We then present the results obtained on the Bacterial Strains dataset. On this dataset, we also employed the two experimental setups of collecting 300 and 600 real samples. However, we further expanded the experimental settings to explore the impact of varying numbers of generated samples. With 300 real samples, we generated 300 and 600 synthetic samples, respectively. With 600 real samples, we generated 600 and 1200 synthetic samples, respectively. The effects of samples generated by different synthetic methods on diagnostic accuracy are shown in Figure 8. The results demonstrate that under these four settings, the data generated by our DiffRaman method yields the most notable enhancement in diagnostic accuracy. With 300 real samples, the use of 300 and 600 synthetic samples can boost the accuracy from 58.1$\%$ to 62.4$\%$ and 64.5$\%$, respectively. In contrast, DA only manages to achieve accuracies ranging from 58.1$\%$ to 58.5$\%$ and 61.4$\%$, cDCGAN from 58.1$\%$ to 60.5$\%$ and 58.4$\%$, and cVAE shows a subtle improvement in accuracy that does not significantly enhance the reliability in diagnostic applications. The same phenomenon is observed when using 600 real samples to generate 600 and 1200 synthetic samples; our method still achieves the best improvement in accuracy, with increases of +5.9$\%$ and 5.7$\%$, outperforming other comparative methods. These results demonstrate that our approach can consistently and effectively improve the diagnostic accuracy of the model under different numbers of synthetic samples generated.

In Figure 9, we present the diagnostic confusion matrices achieved by training diagnostic models with 600 synthetic samples and 300 real experimental samples. It can be observed that our method achieves significant accuracy improvements for some challenging-to-identify categories, such as E. coli and A. baumannii. Furthermore, to more intuitively display the quality of the generated data, we provide in Figure 10 the distribution of data generated by four synthetic methods and test data in the latent space of the last layer of the diagnostic model. It can be observed that the diagnostic model obtained using DiffRaman-generated data exhibits the most discriminative feature differentiation ability, effectively separating different categories of bacteria strains. In contrast, the latent feature distributions obtained by other methods show greater overlap between different categories.

% \begin{table*}[h]
% \centering
% \caption{The accuracy results of spectra synthesized by different generation methods in adapting different diagnostic models}
% \scriptsize
% \begin{tabular}{lcccccccc}
% \toprule
% \multirow{2}{*}{Dataset} & \multirow{2}{*}{Training +Generation Samples} & \multirow{2}{*}{Generation Method}  & \multicolumn{5}{c}{Diagnostic Model} \\
% \cmidrule(lr){4-8}
%  & & & MLP & LSTM & CNN-3 & ResNet & SANet \\
% \midrule
% \multirow{5}{*}{Bacteria ID} & 300 & Source & 0.5 & 0.5 & 0.5 & 0.5 & 0.5 \\
%  & 300+600 & DA & 0.5 & 0.5 & 0.5 & 0.5 & 0.5 \\
%  & 300+600 & cVAE & 0.5 & 0.5 & 0.5 & 0.5 & 0.5 \\
%  & 300+600 & cDCGAN & 0.5 & 0.5 & 0.5 & 0.5 & 0.5 \\
%  & 300+600 & Ours & 0.5 & 0.5 & 0.5 & 0.5 & 0.5 \\
% \midrule
% \multirow{5}{*}{Bacteria Strains} & 600 & Source & 0.5 & 0.5 & 0.5 & 0.5 & 0.5 \\
%  & 600+1200 & DA & 0.5 & 0.5 & 0.5 & 0.5 & 0.5 \\
%  & 600+1200 & cVAE & 0.5 & 0.5 & 0.5 & 0.5 & 0.5 \\
%  & 600+1200 & cDCGAN & 0.5 & 0.5 & 0.5 & 0.5 & 0.5 \\
%  & 600+1200 & Ours & 0.5 & 0.5 & 0.5 & 0.5 & 0.5 \\
% \bottomrule
% \end{tabular}

% \label{table4}
% \end{table*}

To summarize, in this section, we conducted a case study on bacterial identification using Raman spectroscopy. By training the diagnostic model with synthetic bacterial Raman spectral data, we found that it can effectively enhance the accuracy of diagnostic models with limited sample sizes. Additionally, we demonstrated that the proposed DiffRaman method outperforms other methods. Therefore, it has the potential to, for instance, alleviate the manual labor associated with collecting single-cell bacterial Raman spectra and enhance the diagnostic capabilities of intelligent models in clinical scenarios where patient samples are scarce and challenging to gather.

\begin{table*}
\caption{The impact of VQ-VAE's structural hyper-parameters on diagnostic accuracy}
\scriptsize
\begin{tabular}{llllllllll}
\toprule
\multirow{2}{*}{Accuracy} & \multicolumn{3}{l}{d=8} & \multicolumn{3}{l}{d=16} & \multicolumn{3}{l}{d=32} \\ \cline{2-10} 
                          & k=256  & k=512  & k=1024 & k=256  & k=512  & k=1024 & k=256  & k=512  & k=1024  \\ \hline
Bacteria ID                  & 0.510   & 0.542   & 0.550   & 0.586   & 0.593   & \textbf{0.603}   & 0.527   & 0.548   & 0.521    \\
Bacteria Stains            & 0.596   & 0.591   & 0.623   & 0.639   & 0.640   & 0.645   & 0.650   &  \textbf{0.671}   & \underline{0.656}    \\  
AVerage            & 0.553   & 0.567   & 0.587   & 0.612   & \underline{0.616}   & \textbf{0.624}   & 0.588   &  0.609   & 0.588    \\ \bottomrule
\end{tabular}
\end{table*}

\begin{table*}[h]
\centering
\caption{The influence of the diffusion steps in DDPM on diagnostic accuracy}
\resizebox{80mm}{!}
{
\begin{tabular}{lccccc}
\toprule
\multirow{2}{*}{Diffusion Step} & \multicolumn{2}{c}{Bacteria ID} & \multicolumn{2}{c}{Bacteria Stains} \\
\cmidrule(lr){2-3} \cmidrule(lr){4-5}
 & 300+600 & 600+1200 & 300+600 & 600+1200 \\
\midrule
100 & \textbf{0.621} & 0.735 & \textbf{0.671} & \underline{0.784} \\
500 & \underline{0.603} & \textbf{0.756} & \underline{0.645} & \textbf{0.788} \\
1000 & 0.596 & \underline{0.744} & 0.621 & 0.779 \\
\bottomrule
\end{tabular}}
\end{table*}

\begin{table}[h]
\centering
\caption{The comparison between DiffRaman and the original DDPM}
\scriptsize
\resizebox{80mm}{!}
{
\begin{tabular}{lcccc}
\toprule
Method & \makecell{Training \\ Time(s)} & \makecell{Storage \\ Memory(MB)} & \makecell{Bacteria ID\\Accuracy} &  \makecell{Bacteria Stains\\ Accuracy} \\
\midrule
DDPM & 1236 & 266.0  & 0.592 & 0.631 \\
DiffRaman & 787 & 16.5 & 0.603 & 0.645 \\
\bottomrule
\end{tabular}
}
\end{table}
\subsubsection{Further analysis}

In this section, we present additional experimental results. Initially, we explored the hyper-parameter configuration of the two main components of the DiffRaman model. These primarily include the structural hyperparameters of VQ-VAE and DDPM. For VQ-VAE, which is primarily utilized for the latent representation of Raman figure, the key hyper-parameters under investigation include the dimensionality $d$ of the latent feature space and the number $k$ of embedding vectors in the codebook $B$. As shown in Table 4, the impact of different VQ-VAE structures on the accuracy of bacterial isolate identification in the Bacteria ID dataset and species identification in the Bacterial Strains dataset, using 300 real samples and 600 synthetic samples, is presented. The results indicate that the best average diagnostic performance is achieved when using  $d=16$ and $k=1024$. Moreover, with this low-dimensional setting of hyper-parameters, the model is able to maintain good computational efficiency. For DDPM, the most critical hyper-parameter is the number of denoising steps $T$. Similarly, we investigated the impact of 100, 500, and 1000 steps on the accuracy of bacterial strain identification in the Bacterial ID dataset and species identification in the Bacterial Strains dataset. The results are presented in Table 5. Under various configurations of real experimental samples and synthetic samples, the fewer real samples, the better the performance of 100 steps. This is attributed to the fact that the 500-step model struggles to be fully trained with a limited number of real samples. As the dataset size increases, the 500-step model can more accurately fit the data distribution, thus achieving better performance. Based on the above findings, we adopted 500 steps as the default parameter.

Second, we compared DiffRaman with the original DDPM model, which directly generates samples from the raw Raman images. The comparison of computational efficiency and diagnostic performance between the two is shown in Table 6. The results indicate that DiffRaman, by introducing a generation method through latent space, significantly improves computational efficiency, with both the training time and the storage memory required for model storage being substantially reduced. Furthermore, the latent representation of Raman images contains more semantic information, leading to more semantically reasonable generated samples, and thus achieving superior performance (the diagnostic accuracy results were obtained using 300 real samples and 600 synthetic samples).

\section{Conclusions}

In this paper, to address the issue of unstable performance in intelligent Raman spectroscopy diagnostic models trained with limited samples, we introduced DiffRaman, a Raman spectroscopy synthesis framework based on the physically inspired generative model DDPM. We leveraged a limited number of authentic samples to generate a larger quantity of high-quality and realistic synthetic samples, and we trained the diagnostic model using both real and synthetic data to achieve enhanced robustness and generalization capabilities. Validation on two large-scale bacterial Raman spectroscopy datasets has demonstrated that DiffRaman can markedly improve diagnostic accuracy under conditions of limited samples, indicating that the proposed method possesses extensive application potential for alleviating the labor intensity of spectral measurements and in clinical diagnostic scenarios where Raman spectral data of patients are limited.

\section{Acknowledgments}
 This study was supported in part by the Special Project of the Ministry of Industry and Information Technology of China (Grant No. 2023ZY01028).

\printcredits

\bibliographystyle{cas-model2-names}
\bibliography{cas-refs}

\end{document}